\documentclass[letterpaper]{article} 
\usepackage{aaai2026}  
\usepackage{amsmath}
\usepackage{amssymb} 
\usepackage{booktabs}
\usepackage{array}
\usepackage{times}  
\usepackage{helvet}  
\usepackage{courier}  
\usepackage[hyphens]{url}  
\usepackage{graphicx} 
\usepackage{natbib}  
\usepackage{caption} 
\usepackage{algorithm}
\usepackage{algorithmic}

\usepackage{newfloat}
\usepackage{listings}
\DeclareCaptionStyle{ruled}{labelfont=normalfont,labelsep=colon,strut=off} 
\floatstyle{ruled}
\newfloat{listing}{tb}{lst}{}
\floatname{listing}{Listing}
\title{EEGAgent: A Unified Framework for Automated EEG Analysis Using Large Language Models}
\author{
    Sha Zhao\textsuperscript{\rm 1,2}, 
    mingyi peng\textsuperscript{\rm 1,2}, 
    Haiteng Jiang\textsuperscript{\rm 3,4,1}, 
    Tao Li\textsuperscript{\rm 3,4,1}, 
    Shijian Li\textsuperscript{\rm 1,2}
    Gang Pan\textsuperscript{\rm 1,2,4}
}
\affiliations{
    \textsuperscript{\rm 1}State Key Laboratory of Brain-machine Intelligence, Zhejiang University\\
    \textsuperscript{\rm 2}College of Computer Science and Technology, Zhejiang University\\
    \textsuperscript{\rm 3}Department of Neurobiology, Affiliated Mental Health Center \& Hangzhou\\
    Seventh People’s Hospital, Zhejiang University School of Medicine\\
    \textsuperscript{\rm 4}MOE Frontier Science Center for Brain Science and Brain-machine Integration,
    Zhejiang University\\
}

\usepackage{bibentry}

\begin{document}
	
\maketitle

\begin{abstract}
Scalable and generalizable analysis of brain activity is essential for advancing both clinical diagnostics and cognitive research. Electroencephalography (EEG), a non-invasive modality with high temporal resolution, has been widely used for brain states analysis. However, most existing EEG models are usually tailored for individual specific tasks, limiting their utility in realistic scenarios where EEG analysis often involves multi-task and continuous reasoning. In this work, we introduce EEGAgent, a general-purpose framework that leverages large language models (LLMs) to schedule and plan multiple tools to automatically complete EEG-related tasks. EEGAgent is capable of performing the key functions: EEG basic information perception, spatiotemporal EEG exploration, EEG event detection, interaction with users, and EEG report generation. To realize these capabilities, we design a toolbox composed of different tools for EEG preprocessing, feature extraction, event detection, etc. These capabilities were evaluated on public datasets, and our EEGAgent can support flexible and interpretable EEG analysis, highlighting its potential for real-world clinical applications.
\end{abstract}

%

\section{Introduction}
\label{sec:introduction}
Brain diseases are increasingly recognized as a major global health challenge, often leading to profound functional impairments and progressive cognitive decline \cite{steinmetz2024global}. To support early diagnosis and effective intervention, the ability to accurately and efficiently assess brain function is of paramount importance. Among existing neurophysiological techniques, electroencephalography (EEG) stands out as a widely accessible, non-invasive tool that offers high temporal resolution and low operational cost. By capturing electrical signals generated by neuronal activity via scalp electrodes, EEG provides rich, multi-channel time-series data that encode dynamic information about both physiological processes and pathological changes. Despite its value, EEG analysis presents substantial challenges. The signals are highly non-stationary, exhibit considerable inter-subject variability, and contain complex waveform patterns that are difficult to interpret \cite{Shen2019}. Manual analysis by domain experts remains the standard in clinical practice, but this approach is time-consuming, labor-intensive, and inherently subjective. \textbf{These constraints severely limit the scalability and efficiency of EEG applications, especially in settings requiring continuous monitoring or multi-faceted analysis.}

Recent advances in machine learning (ML) and deep learning (DL) have enabled more automated and scalable approaches to EEG analysis\cite{zhou2025spiced}. Modern data-driven models can learn meaningful representations directly from raw EEG signals and achieve high accuracy across tasks such as seizure detection, sleep staging, and neurological disorder diagnosis \cite{li2025}. These approaches have led to notable improvements in analysis speed, accuracy, and reproducibility. \textbf{However, most existing EEG models remain task-specific and are optimized for isolated objectives.} This leads to the so-called task isolation problem, wherein each model addresses only a narrow slice of EEG analysis without considering the broader context or the interdependence of concurrent tasks. In clinical scenarios, EEG analysis often involves multi-task, continuous reasoning—for example, detecting epileptiform discharges while also assessing cognitive states or sleep transitions \cite{zhao2024systematic}. \textbf{Most current task-specific EEG models fail to meet this requirement, necessitating an urgent development of a unified and flexible framework to address this critical need.} Fortunately, recent breakthroughs in large language models (LLMs) offer a promising pathway toward overcoming these limitations \cite{zhang2024}. With their powerful capabilities in task planning, contextual reasoning, and tool integration, LLM-powered agents can manage complex, multi-step workflows. They dynamically interpret user intent and interface with external modules, making them ideal candidates for scheduling multi-task EEG analysis in a coordinated and intelligent manner. 

Inspired by this potential, we propose \textbf{EEGAgent}, a general-purpose, LLM-enhanced intelligent framework for EEG analysis. \textbf{EEGAgent is designed to bridge the gap between single-task models and multiple  EEG analysis demands in the real world.} It achieves this by integrating perception, task understanding, and dynamic model scheduling into a unified system. EEGAgent brings together traditional feature engineering, multiple deep learning backbones, and an LLM-based policy planning engine to enable end-to-end coordination and execution of diverse EEG tasks. Through this integration, EEGAgent performs modular and adaptive scheduling, dynamically selecting appropriate pipelines based on the characteristics of the input data and the goals of the analysis. Furthermore, the system is equipped with domain-specific knowledge bases and the ability to interface with external information sources, enhancing its contextual awareness and interpretability. With this architecture, EEGAgent supports the entire workflow, from signal preprocessing and feature extraction to classification, decision reasoning, and interactive reporting. By providing an intelligent, multi-task-capable EEG analysis platform, our framework aims to advance the field toward greater automation, adaptability, and clinical readiness, helping unlock the full potential of EEG as a diagnostic and monitoring modality. Our key contributions are as follows:
\begin{itemize}
	\item We propose the EEGAgent framework, which enables unified scheduling and automated execution of multi-task EEG analysis by integrating traditional and deep learning-based methods, thereby improving overall system efficiency and adaptability. To our best knowledge, we are the first to design an agent for EEG analysis.
	\item EEGAgent incorporates key EEG-specific capabilities such as context awareness, flexible spatiotemporal analysis, accurate event localization, and automated interactive reporting, facilitating comprehensive and adaptive EEG analysis and interpretation.
	\item We evaluate the capability of EEGAgent on public datasets, demonstrating its clinical potential through the support for interactive workflows and automated report generation, which can enhance diagnostic efficiency and assist clinical decision-making.
\end{itemize}

\section{Related works}
\subsection{EEG Analysis}
Electroencephalography (EEG) is widely used in sleep staging, emotion recognition, seizure detection, and cognitive workload assessment, all of which require models capable of extracting informative patterns from noisy brain signals. Existing methods fall into two major categories: traditional machine learning with handcrafted features and deep learning models that learn directly from raw data.

Traditional ML approaches remain useful where efficiency or interpretability is important. Naive Bayes has been applied to seizure and drowsiness detection \cite{Ashok2016}, Random Forests to mental workload and emotion recognition \cite{messaoud2021random}, and Linear Discriminant Analysis (LDA) to motor imagery–based BCIs \cite{Fu2020, Santos2023}. Other classical classifiers such as SVM, KNN, and Logistic Regression are frequently used as baselines or in hybrid systems \cite{saeidi2021neural}.
Deep learning methods have shown strong performance by automatically extracting spatial–temporal features from EEG signals \cite{zhoubrainuicl,Wang_Zhao_Jiang_Li_Li_Pan_2024}. CNNs excel in sleep staging and motor imagery \cite{efe2023cosleepnet, khademi2022transfer}, RNNs (e.g., LSTM, GRU) perform well in sequential tasks such as seizure and emotion detection \cite{mekruksavanich2023effective, 10.1145/3746027.3755270, WANG2025130008}, and Transformer-based models further improve the modeling of long-range dependencies \cite{Sun2021, wei2023}.
However, most existing EEG models are designed for isolated tasks under controlled conditions, which limits their practical applicability. In real-world scenarios, tasks such as artifact removal, event detection, and signal classification often need to be performed together, underscoring the need for a unified, context-aware framework that can flexibly support multi-task EEG analysis.

\begin{figure*}[htbp] 
	\centering 
	\includegraphics[width=\textwidth]{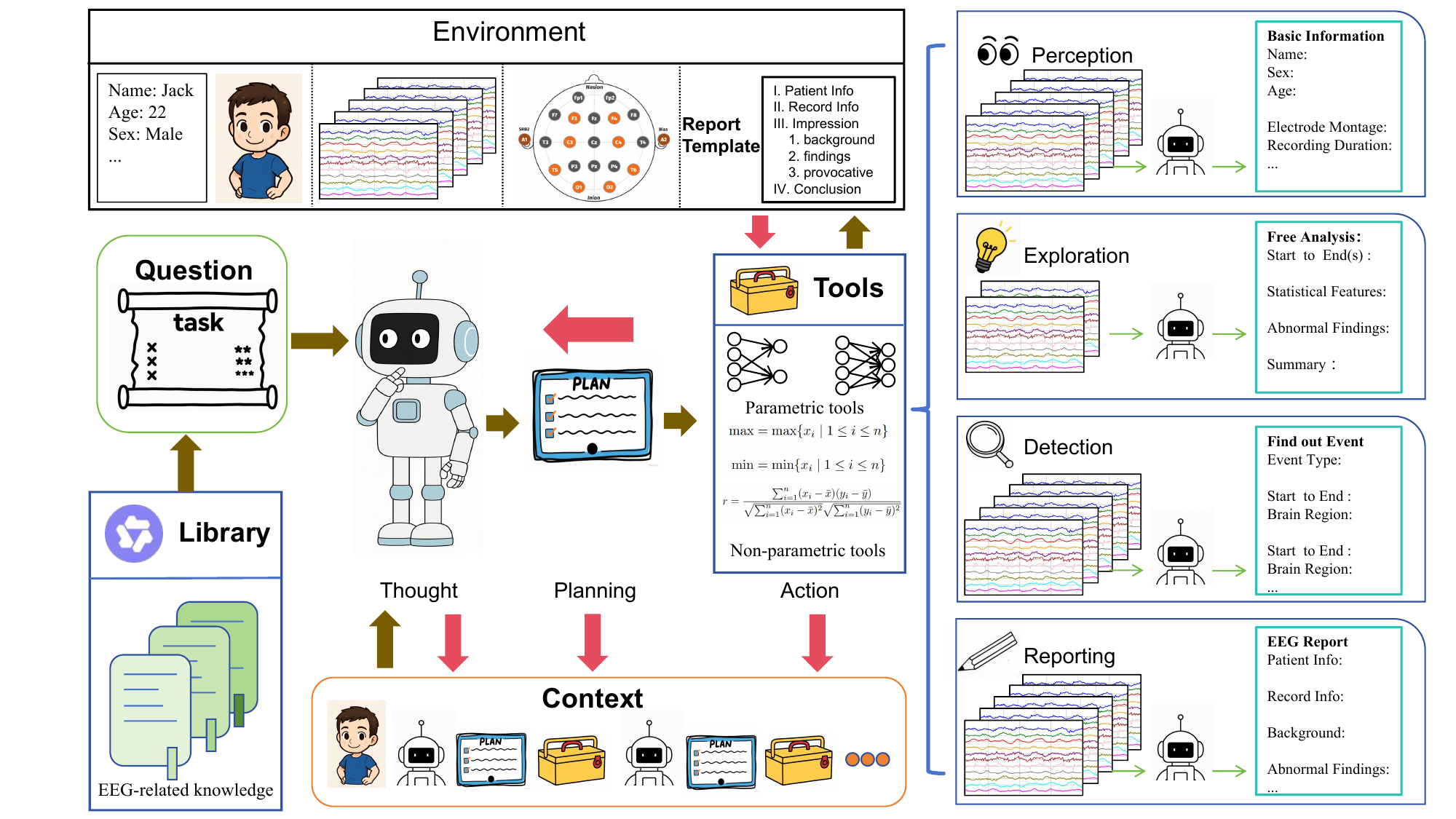} 
	\caption{EEGAgent framework} 
	\label{fig:framework}
\end{figure*}

\subsection{LLM-Enhanced Agent Framework}
Recent advances in large language models (LLMs) have enabled the creation of intelligent agents that can perform complex reasoning and manage multi-step tasks\cite{wang2025cbramodcrisscrossbrainfoundation}. Originally designed for next-token prediction, LLMs now demonstrate impressive capabilities in language understanding, planning, and decision-making \cite{brown2020language, radford2019LanguageMA}. When augmented with modules for planning, memory, perception, and tool use, they can coordinate workflows, invoke external tools, and synthesize information to support high-level decision-making \cite{yao2023react, shinn2023reflexion}. Techniques such as Chain of Thought (CoT) \cite{wei2022COT} and Tree of Thought (ToT) \cite{yao2023TOT} enhance LLM reasoning via stepwise decomposition and hypothesis exploration. Planning modules help break down goals into subtasks, while memory mechanisms enable both contextual tracking and long-term knowledge retention \cite{huang2022language, wang2023voyager, wu2022memorizingtransformers}. Retrieval-Augmented Generation (RAG) further improves reasoning by allowing access to external knowledge bases such as clinical guidelines \cite{lewis2020retrieval}.

However, directly applying LLM agents to EEG analysis poses unique challenges. EEG signals are highly variable and non-stationary, and they require domain-specific interpretation. Tailoring LLMs to accommodate these complexities is essential to ensure their utility in EEG-related applications. While most existing EEG models remain task-specific, recent trends in LLM-driven agent frameworks offer a new direction. These agents leverage planning, memory, and tool integration to coordinate analytic workflows and generate structured outputs. Extending such capabilities to EEG analysis could enable more flexible, multi-task solutions that align with clinical workflows.

\section{Method}
The EEGAgent is an intelligent system designed to autonomously plan and execute EEG analysis tasks. As illustrated in Fig.~\ref{fig:framework}, this framework leverages a series of tools that automate the execution of these tasks, thereby forming several distinctive capabilities tailored to diverse EEG analysis scenarios.
Specifically, the agent demonstrates the following core competencies:
\begin{itemize}
	\item \textbf{Perception}: The EEGAgent is aware of the EEG data context, including subject metadata, available channels, and the temporal span of the recording.
	\item \textbf{Exploration}: The EEGAgent supports in-depth analysis within user-specified or system-identified spatiotemporal ranges. It flexibly schedules appropriate tools—such as frequency analysis, waveform extraction, or statistical characterization—to examine targeted EEG segments from multiple perspectives.
	\item \textbf{Detection}: The EEGAgent can accurately identify the time segments and brain regions where specific EEG events occur and further decode brain states, enabling efficient retrieval and matching.
	\item \textbf{Interaction and Reporting}: The EEGAgent supports interaction with users and automatically generates structured EEG reports based on predefined templates.
\end{itemize}

To support these capabilities, the EEGAgent integrates a modular toolbox of parametric models and statistical feature extractors. Guided by a large language model, it dynamically plans and sequences tool usage based on task needs and context. This design enables flexible, context-aware analysis—from broad perception to precise localization—along with automated and structured report generation.

\newcommand{\thicktoprule}{\specialrule{1.5pt}{0pt}{0pt}}    
\newcommand{\thickmidrule}{\specialrule{1.2pt}{0pt}{0pt}}    
\newcommand{\thickbottomrule}{\specialrule{1.5pt}{0pt}{0pt}} 

\begin{table*}[h!]
	\centering
	\small  
	\resizebox{\textwidth}{!}{
		\begin{tabular}{
				>{\centering\arraybackslash}m{3cm}
				>{\centering\arraybackslash}m{2.5cm}
				>{\centering\arraybackslash}m{3cm}
				>{\centering\arraybackslash}m{3cm}
				m{7cm}
			}
			\thicktoprule
			\textbf{Tool Name} & \textbf{Type} & \textbf{Time Granularity} & \textbf{Space Granularity} & \textbf{Description} \\
			\thickmidrule
			\texttt{normalAbnormal} & Parametric Tool & Full EEG & Whole Channel & Estimates probability of pathological normality/abnormality for the entire EEG. \\
			\midrule
			\texttt{eyemMuscle} & Parametric Tool & Single Second & Single Channel & Classifies eye movement and muscle artifacts within 1-second window per channel. \\
			\midrule
			\texttt{seizArtiBckg} & Parametric Tool & Single Second & Single Channel & Classifies seizure, artifact, and background within 1-second window per channel. \\
			\midrule
			\texttt{seizNormal} & Parametric Tool & Single Second & Single Channel & Detects seizure vs. non-seizure within 1-second window per channel. \\
			\midrule
			\texttt{slowSeizBckg} & Parametric Tool & 10 Seconds & Whole Channel & Classifies slow waves, epileptic and background activity in 10-second windows. \\
			\midrule
			\texttt{baseInfo} & Non-parametric Tool & Full EEG & Whole Channel & Extract structured information such as patient demographics and recording metadata from the EEG. \\
			\midrule
			\texttt{compute\_amplitude} & Non-parametric Tool & $\leq$ 60 Seconds & Whole Channel & Computes amplitude features (mean abs, RMS, max/min) for selected channels. \\
			\midrule
			\texttt{compute\_psd} & Non-parametric Tool & $\leq$ 60 Seconds & Whole Channel & Calculates power spectral density across frequency bands for selected channels. \\
			\midrule
			\texttt{compute\_symmetry} & Non-parametric Tool & $\leq$ 60 Seconds & Left-Right Channel Pair & Calculates Pearson correlation for left-right channel pairs to assess symmetry. \\
			\thickbottomrule
		\end{tabular}
	}
	\caption{Tools and their time-space granularity.}
	\label{table:tools_time_space}
\end{table*}

\subsection{EEGAgent Architecture}
The EEGAgent is designed to operate in a complex environment consisting of subject information, EEG recording metadata, raw signals, and report templates. It is built on Qwen3-235B ~\cite{yang2025qwen3technicalreport}, a large language model known for its strong reasoning capabilities and seamless integration with external tools, hereafter referred to as Qwen. 
To provide domain-specific expertise, the Agent incorporates an EEG knowledge base. When a new task arrives, the task description is embedded into a high-dimensional semantic vector using Qwen3-Embedding-8B~\cite{zhang2025qwen3embedding}, and relevant knowledge entries are retrieved from the knowledge base via similarity search. The retrieved information is then used to enrich the context for Qwen. 
The Agent’s toolbox contains two types of tools: parametric tools based on deep learning models and non-parametric tools that utilize handcrafted statistical features. Upon receiving a task, Qwen combines the enriched context and the current EEG data to analyze the problem, formulate a plan, and select the appropriate tools. These tools are then executed to extract the necessary information, which is fed back to Qwen for further reasoning. This iterative cycle continues until the task is satisfactorily completed. 
\textbf{By integrating knowledge retrieval, memory, and a versatile toolbox under the control of Qwen, the EEGAgent provides a flexible and interpretable framework capable of handling diverse EEG analysis tasks in an automated manner.}

\subsection{Toolbox Design for Capability Support}
To realize the core capabilities of the EEGAgent, the toolbox is organized along three independent dimensions: parametric vs. non-parametric nature, temporal granularity, and spatial granularity. These dimensions characterize each tool’s functional attributes and computational cost, enabling flexible scheduling and intelligent composition. Parametric deep learning tools capture high-level clinical semantics for complex tasks, whereas non-parametric statistical and signal-processing tools provide efficient extraction of fundamental features. Temporal granularity ranges from coarse long-window analysis to fine short-segment detection, and spatial granularity spans global multi-channel assessment to single-channel localization.

The Agent flexibly schedules and combines these tools to meet task requirements. For instance, non-parametric tools such as \texttt{baseInfo} support macro-level perception, while parametric tools such as \texttt{seizNormal} and \texttt{slowSeizBckg} enable precise event localization. By dynamically integrating tools of different natures and granularities, the Agent achieves multi-perspective analyses and balances accuracy with efficiency. The resulting multidimensional information is further fused to drive interaction and report generation, ultimately producing structured and clinically meaningful EEG outputs. A summary of the tools and their temporal–spatial characteristics is shown in Table~\ref{table:tools_time_space}.

\subsection{EEG Perception Ability: Comprehending the fundamental information of EEG signals} Perception capability refers to that the EEGAgent can understand the EEG data environment in which it operates. Given the specific characteristics of EEG data, the EEGAgent can not only access basic information but also extract critical advanced features, enabling it to build a comprehensive perceptual model of EEG recordings. EEG data inherently contains rich contextual information, such as electrode configuration, recording duration, and subject status. To enhance the Agent’s understanding of such background details, \textbf{we developed a compact knowledge module encapsulated as the baseInfo tool, which extracts key metadata upon EEG loading.} This tool is automatically triggered during the initial loading phase (i.e., environment initialization), helping the Agent to construct a foundational perceptual view.

Within this module, the system extracts essential subject information (e.g., name, gender, age), the total recording duration, and the electrode montage. Among these, age plays a particularly crucial role in EEG interpretation. To improve the Agent’s sensitivity to age-related factors, we incorporate summarized insights from clinical literature regarding normative EEG patterns across developmental stages \cite{Liu2011}, allowing the Agent to apply age-appropriate reference standards. Besides, the spatial organization of electrodes and their mapping to anatomical brain regions are presented in a structured tabular format, enabling the Agent to understand spatial relationships among channels, including symmetry and brain regional affiliation. We also provide a concise summary of typical frequency bands and their associated pathological features to serve as cognitive references. Since foundational information such as age-related factors and electrode configurations is independent of specific tasks, it can be preloaded before any analysis begins. \textbf{This enables the EEGAgent to form an initial expectation of EEG patterns}, effectively simulating the anticipatory cognition that human experts develop prior to interpreting EEG signals.

\subsection{EEG Exploration Ability: Analyzing  EEG segments from multiple perspectives}
The EEGAgent’s exploration ability enables it to conduct flexible and systematic analysis of any task-specified time interval within EEG recordings.
To model the EEGAgent’s exploration capability, we consider a  temporal interval \([T_{\text{start}}, T_{\text{end}}]\), which is partitioned into \(N\) non-overlapping segments of length \(\Delta t\):
\begin{equation}
	X_i = [T_{\text{start}} + (i{-}1)\Delta t,\; T_{\text{start}} + i\Delta t], \quad i = 1, \dots, N.
\end{equation}
For each segment \(X_i\), the Agent selects an analysis plan \(\mathcal{T}_i\) from the tool library via a Qwen-based controller \(\pi_\theta\), conditioned on contextual information \(C\) (such as patient age, prior events, modality priors):
\begin{equation}
	\mathcal{T}_i = \pi_\theta(X_i,\ C) \subseteq \mathcal{T}_{\mathrm{p}} \cup \mathcal{T}_{\mathrm{np}},
\end{equation}
where \(\mathcal{T}_{\mathrm{p}}\) and \(\mathcal{T}_{\mathrm{np}}\) represent sets of parametric and non-parametric tools, respectively.
Each tool \(T \in \mathcal{T}_i\) is treated as a function:
\begin{equation}
	T: X_i \mapsto R_i^T, \quad R_i^T \in \mathbb{R}^{d_T} \ \text{or} \ \mathcal{V}^*,
\end{equation}
where \(R_i^T\) denotes the tool's output, such as feature vectors, classification scores, or symbolic descriptors.
To synthesize tool outputs for each segment, a fusion function \(f_{\mathrm{fuse}}\) integrates intermediate results:
\begin{equation}
	R_i = f_{\mathrm{fuse}}(\{R_i^T\};\ \phi), \quad \phi = \text{fusion parameters}.
\end{equation}
Finally, a summarization function \(f_{\mathrm{summarize}}\) aggregates results across all segments into a comprehensive semantic summary \(\mathcal{S}\):
\begin{equation}
	\mathcal{S} = f_{\mathrm{summarize}}(\{R_i\}_{i=1}^{N};\ C),
\end{equation}
which includes global assessments, rhythmic patterns, and localized event characterizations. The entire exploration workflow is coordinated by a large language model. It handles every step, from selecting tools to fusing outputs and generating summaries. Rather than relying on fixed rules, the Agent interprets tool results based on context. This approach resolves representational mismatches and allows the system to generate high-level insights through language-based reasoning. As a result, the EEGAgent can explore EEG data like a human expert, with a balance between interpretability and efficiency across different analysis goals.

\subsection{EEG Detection Ability: Recognizing Targeted Spatiotemporal EEG Event}
The EEGAgent is capable of automatically identifying specific clinically relevant EEG event types during review, and precisely localizing these events in both temporal and spatial domains throughout the recording. Given the non-stationary nature of EEG signals and the dynamic variation of their statistical properties over time, analyses conducted at a single temporal scale are insufficient to accurately capture event features. To address this challenge, \textbf{we adopt a multi-granularity analysis strategy} that improves detection accuracy and efficiency while reasonably controlling computational costs. In the temporal dimension, the EEGAgent utilizes two time scales: 10 seconds and 1 second. The 10-second window corresponds to the typical duration used in clinical review and serves as a coarse screening tool to determine whether a target event occurs within that period. Upon detection of a potential event, the EEGAgent further switches to a finer 1-second scale to conduct detailed analysis, achieving more precise spatiotemporal localization. In the spatial dimension, the EEGAgent employs both single-channel and multi-channel analyses, enabling it to focus on local signal variations at individual electrodes as well as integrate information across multiple channels to enhance robustness in event detection.

Notably, the entire analysis process is based on a sliding window mechanism and is dynamically planned and scheduled by Qwen. It adjusts the analysis strategy according to task requirements and intermediate results, balancing the trade-off between analysis granularity and computational cost. This flexible guidance enables the EEGAgent to progressively refine its analysis from coarse to fine granularity. As a result, event detection and localization are performed efficiently and economically without compromising accuracy.

\begin{algorithm}[H]
	\caption{EEG Report Generation}
	\label{alg:eeg_report}
	\textbf{Input}: Raw EEG data $D$ \\
	\textbf{Parameter}: Tool list $\mathcal{L}$ \\
	\textbf{Output}: Final EEG report $R$
	\begin{algorithmic}[1]
		
		\STATE Initialize report template $\mathcal{T}$
		\STATE $\mathcal{S} \leftarrow \text{Segment}(D,\ 10s)$ \COMMENT{Divide EEG into 10s windows}
		
		\FOR{each segment $s_i \in \mathcal{S}$}
		\STATE $r_i^{\text{coarse}} \leftarrow \text{Analyze}(s_i,\ \texttt{coarse})$ \COMMENT{Coarse-level analysis}
		\IF{$\text{Qwen\_Decide}(r_i^{\text{coarse}}) = \texttt{fine}$}
		\FOR{each subsegment $f_{i,j} \in \text{Segment}(s_i,\ 1s)$}
		\STATE $r_{i,j}^{\text{fine}} \leftarrow \text{Analyze}(f_{i,j},\ \texttt{fine})$
		\STATE $\mathcal{T} \leftarrow \mathcal{T} \cup r_{i,j}^{\text{fine}}$
		\ENDFOR
		\ELSE
		\STATE $\mathcal{T} \leftarrow \mathcal{T} \cup r_i^{\text{coarse}}$
		\ENDIF
		\ENDFOR
		
		\STATE $R \leftarrow \text{GenerateReport}(\mathcal{T})$ \COMMENT{Qwen or template-based synthesis}
		\RETURN $R$
	\end{algorithmic}
\end{algorithm}

\subsection{Interaction and Reporting Ability: Interacting with Users and Generating Clinical EEG Reports}
EEGAgent enables lightweight user interaction by remembering context, allowing users to ask follow-up questions, request specific analyses, or clarify ambiguous results. This conversational ability improves interpretability and aligns with clinical workflows. Building on this interaction, EEGAgent can automatically generate structured reports to assist neurologists, summarizing both global patterns and local abnormalities through a hierarchical, multi-scale analysis pipeline. To support this goal, EEGAgent adopts a modular reporting strategy based on the ACNS standardized template \cite{tatum2016acns}, decomposing the report into components such as background activity, abnormal events, and clinical impressions. This design ensures consistency and interpretability while accommodating variability across clinical cases. As shown in Algorithm~\ref{alg:eeg_report}, the input EEG is first segmented into 10-second windows, aligned with standard clinical practice. Each segment is analyzed coarsely to identify salient events. A lightweight Qwen controller determines whether finer 1-second-scale analysis is warranted, allowing the system to focus on diagnostically relevant regions without overreacting to noise or isolated outliers. The resulting observations from both granularities are compiled into a structured template and converted into final text using either template-based generation or Qwen-assisted summarization. This unified workflow enables EEGAgent to produce clinically aligned reports that are robust, interpretable, and capable of supporting multi-level reasoning over complex EEG signals.

\section{Experiments}
\subsection{Dataset}
We evaluate the EEGAgent on three widely used benchmark datasets from the Temple University Hospital (TUH) EEG corpus. 
The TUH EEG corpus includes several subsets, each targeting distinct analysis tasks and annotation granularities.  
\textbf{TUH Abnormal EEG Corpus (TUAB)} \cite{López2015} is designed for binary classification of EEG recordings as normal or abnormal. 
It contains 2,993 recordings from 2,383 unique subjects, with roughly equal proportions of 1,521 normal and 1,472 abnormal samples. 
Each recording spans approximately 20 minutes and follows the standard 10–20 electrode montage, with a typical sampling rate of 250~Hz. 
Annotations are provided at the recording level, making TUAB suitable for global assessment.  
\textbf{TUH EEG Event Corpus (TUEV)} \cite{Obeid2016} focuses on fine-grained event classification. 
It comprises 518 recordings annotated with six event types: periodic lateralized epileptiform discharges (PLED), generalized periodic discharges (GPED), spikes and sharp waves (SPSW), eye movements (EYEM), artifacts (ARTF), and background activity (BCKG). 
Annotations are provided at 1-second resolution on a per-channel basis, enabling precise temporal and spatial localization. 
TUEV supports various classification tasks, including binary seizure detection, artifact discrimination, and multi-class event recognition. 
We evaluate the detection capability of the EEGAgent on this dataset.  
\textbf{TUH EEG Slowing Corpus (TUSL)} \cite{Obeid2016} targets the differentiation of epileptiform slowing from background activity and seizure events. 
It includes 300 labeled segments, each 10 seconds long, from 75 recording sessions involving 38 patients. 
All data are recorded with the TCP-REF montage at 256~Hz. 
Annotations are provided at the segment level across all channels. 
Due to its limited size, ensemble strategies such as soft voting are often used to improve evaluation robustness.  
Together, these datasets provide a comprehensive testbed for multi-scale, multi-task EEG analysis, supporting both global and localized objectives under varying levels of annotation granularity.

\begin{figure}[htbp] 
	\centering 
	\includegraphics[width=\columnwidth]{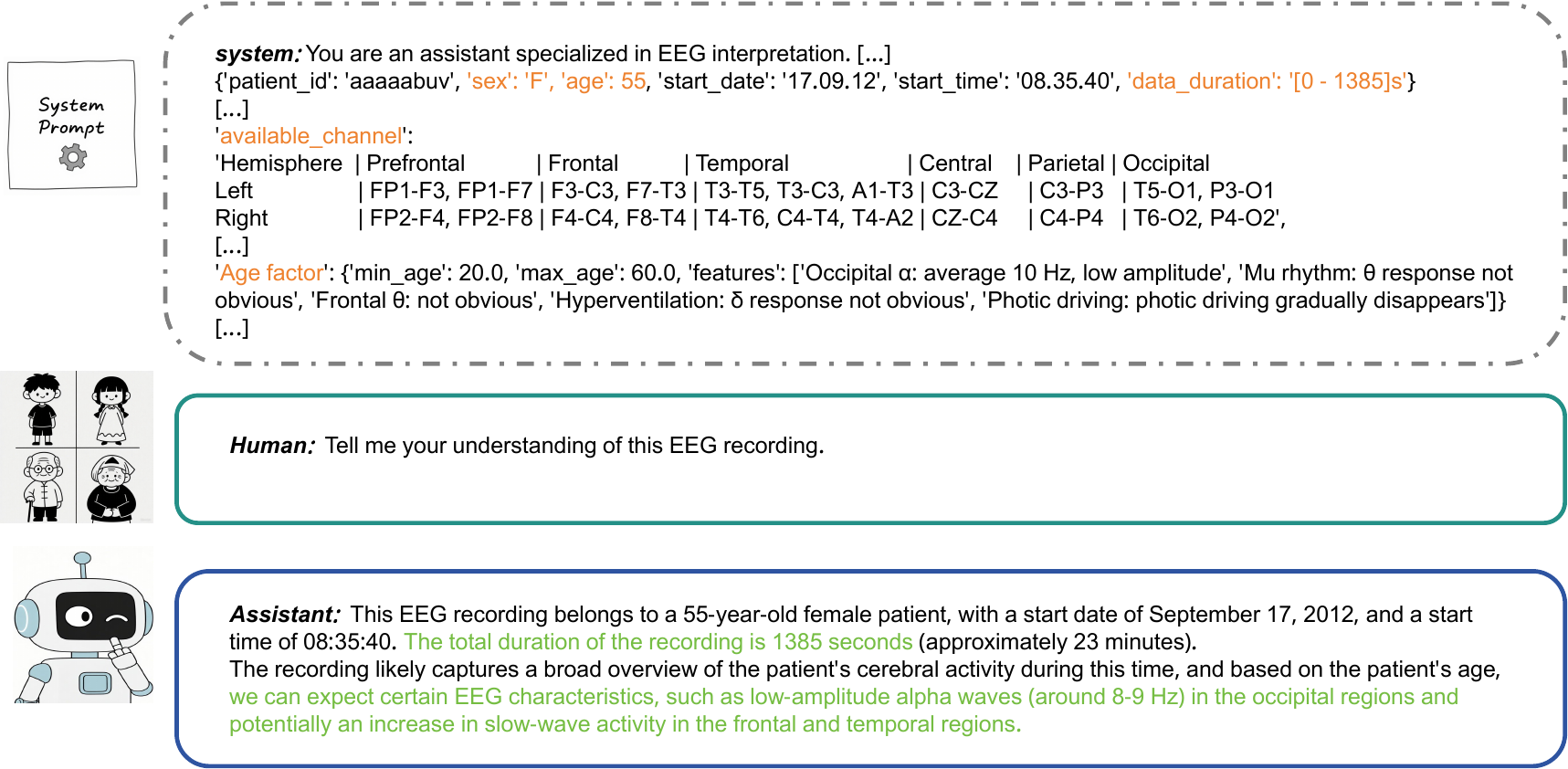} 
	\caption{The EEGAgent perceives the environment constructed from EEG data. Solid lines represent processes visible to the user.}
	\label{fig:exp1_}
\end{figure}

\subsection{EEG Perception Experiment}
We evaluate the perceptual capability of the EEGAgent on TUAB. This experiment is designed to assess whether the Agent can accurately extract contextual metadata and integrate prior neurophysiological knowledge into its interpretation process. As shown in Figure~\ref{fig:exp1_}, the Agent is provided with structured information, including patient demographics (e.g., sex and age), recording parameters (e.g., start date, start time, and total duration), and spatial channel mappings (e.g., FP1-F3 and F3-C3 corresponding to the left frontal region).

The evaluation focuses on the Agent’s ability to extract key metadata such as the subject’s age and the recording duration, and subsequently apply these contextual cues when interpreting the EEG background activity. In this case, the patient’s age informs the Agent’s expectations regarding typical EEG characteristics, including slowing of background rhythms, reduced fast activity, the presence of low-amplitude occipital alpha rhythms, and a possible increase in frontal and temporal slow-wave activity. This reasoning process leverages well-established age-related EEG patterns \cite{Liu2011}. The objective of this evaluation is to determine whether the EEGAgent can demonstrate context-driven perceptual capability by combining demographic information, spatial channel knowledge, and neurophysiological priors to guide its understanding of the EEG environment.

\begin{figure}[htbp] 
	\centering 
	\includegraphics[width=\columnwidth]{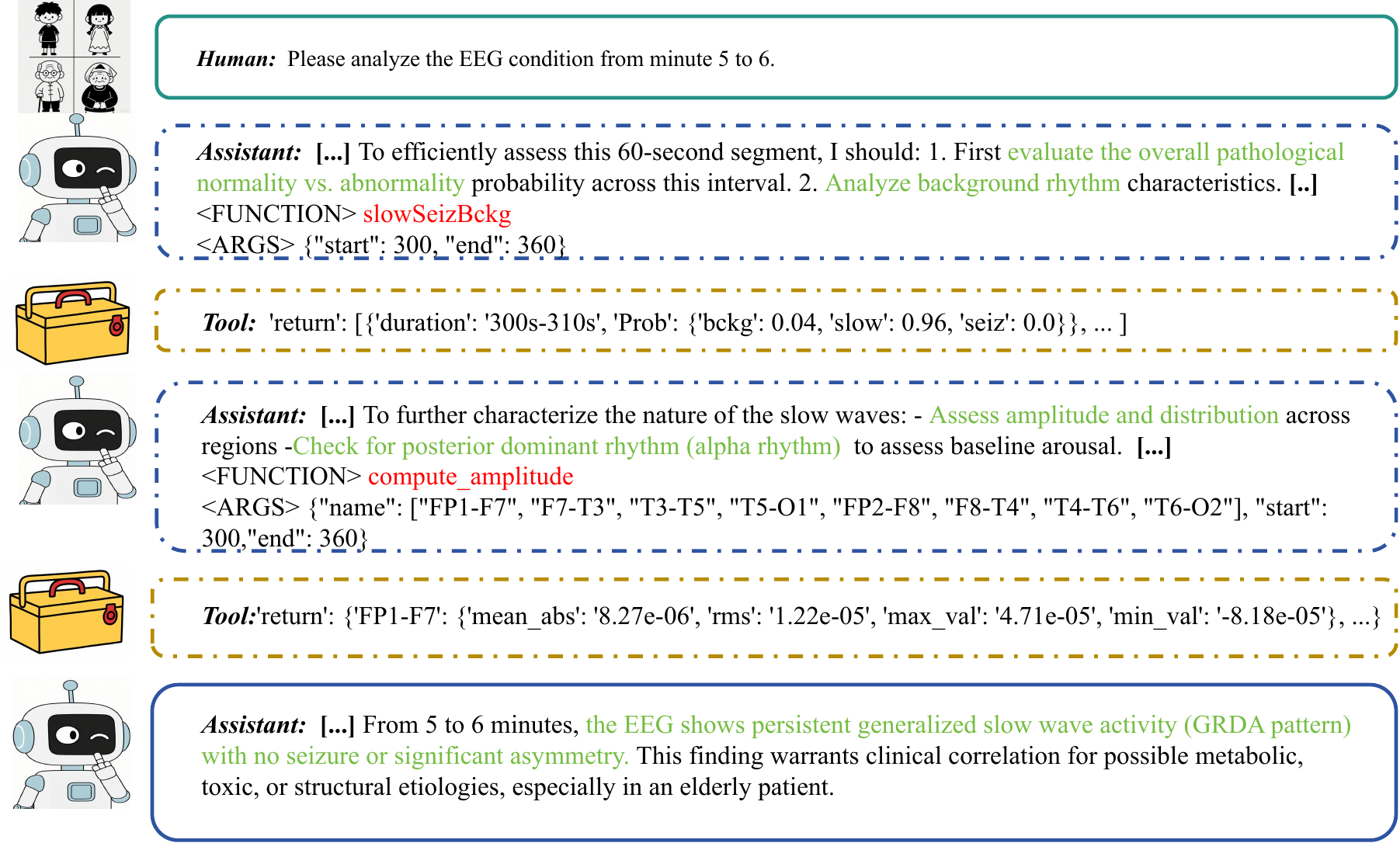} 
	\caption{Exploration of EEG segments by the EEGAgent. Solid lines represent processes visible to the user.} 
	\label{fig:exp3_}
\end{figure}

\subsection{EEG Segment Exploration Experiment}
To evaluate the Agent’s exploratory capability, we present a case study from the TUAB corpus in response to the non-specific query: “Please analyze the EEG condition from minute 5 to 6.” As shown in Figure~\ref{fig:exp3_}, the Agent autonomously executes a structured, multi-stage workflow. First, it deploys the parametric tool \texttt{slowSeizBckg} for a rapid assessment, which detects slow-wave activity with a high probability. Based on this preliminary result, the Agent initiates a deeper analysis using the non-parametric tool \texttt{compute\_amplitude} to precisely characterize the spatial distribution of the detected waves. In the final stage, the Agent synthesizes multiple sources of evidence: the high-level classification (\texttt{slow wave}), low-level quantitative features (indicating a generalized pattern), and relevant patient context (\texttt{elderly}). This integrative reasoning produces a clinically meaningful report identifying \emph{persistent generalized slow-wave activity}, which warrants further clinical correlation. This case study \textbf{highlights the Agent’s ability to schedule a coarse-to-fine sequence of analytical tools}, effectively bridging the gap between raw EEG data and expert-level interpretation.

\begin{figure}[htbp] 
	\centering 
	\includegraphics[width=\columnwidth]{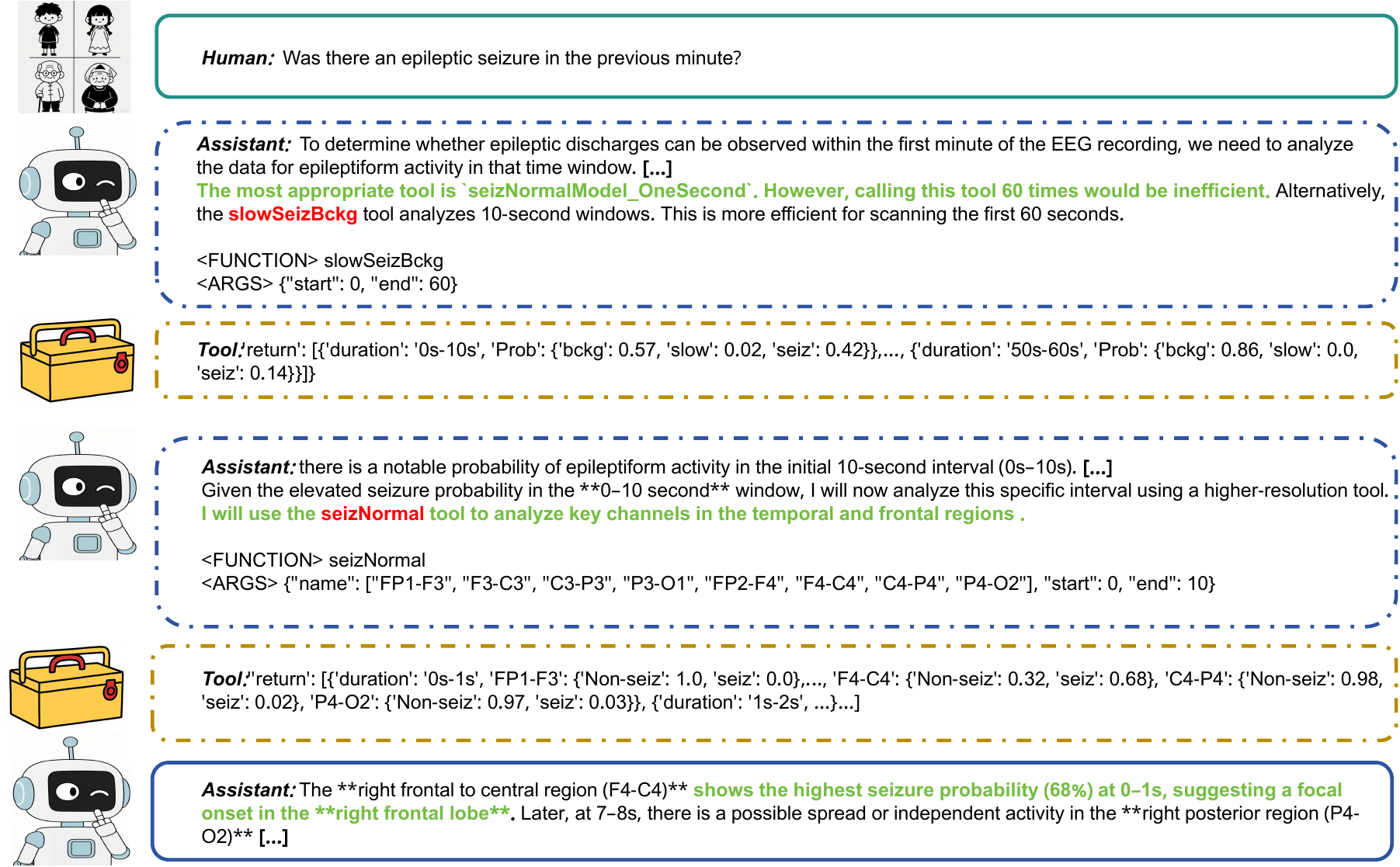} 
	\caption{Detection of Epileptic Discharges by the EEGAgent. Solid lines represent processes visible to the user.} 
	\label{fig:exp2_}
\end{figure}

\subsection{EEG Event Detection Experiment}
Detection ability involves not only identifying the temporal intervals of EEG events but also localizing their spatial distribution. 
Because many clinically relevant events are transient and spatially variable, EEGAgent adopts a hierarchical multi-scale strategy to balance accuracy and efficiency. 
As shown in Figure~\ref{fig:exp2_}, the system first performs a coarse-grained scan (e.g., 60-s windows) to quickly locate suspicious intervals, and then progressively focuses on these intervals using higher-resolution tools to refine temporal and spatial localization. 
In the example, an initial elevated seizure probability at 0–10 s was further resolved at the channel level, ultimately pinpointing the right fronto-central region (F4–C4) at 0–1 s as the likely onset site.

Additionally, we evaluated this detection framework on the TUEV test set, focusing on a seizure detection task. 
Annotations were preprocessed by merging adjacent events less than 1 s apart and defining SPSW, GPED, and PLED as positive classes. 
Predictions were considered correct when their IoU with the reference events exceeded 0.7 on the same channel. 
Under this criterion, the framework achieved a hit rate of 69.30\% and a false rate of 44.77\%, demonstrating that the stepwise localization strategy enables efficient and precise detection of transient epileptiform events.

\subsection{EEG Report Generation Experiment}
We conducted a feasibility evaluation of the EEGAgent’s expression capability using the TUAB dataset, a large-scale clinical EEG dataset with normal and abnormal labels that reflects the information distribution typically encountered in real-world clinical settings. 
Due to the lack of paired EEG recordings and standardized reports, \textbf{we designed an automatic report generation task} based on ACNS standards \cite{tatum2016acns} to assess the system’s ability to translate signal analysis results into clinical language.

\begin{figure}[htbp] 
	\centering 
	\includegraphics[width=\columnwidth]{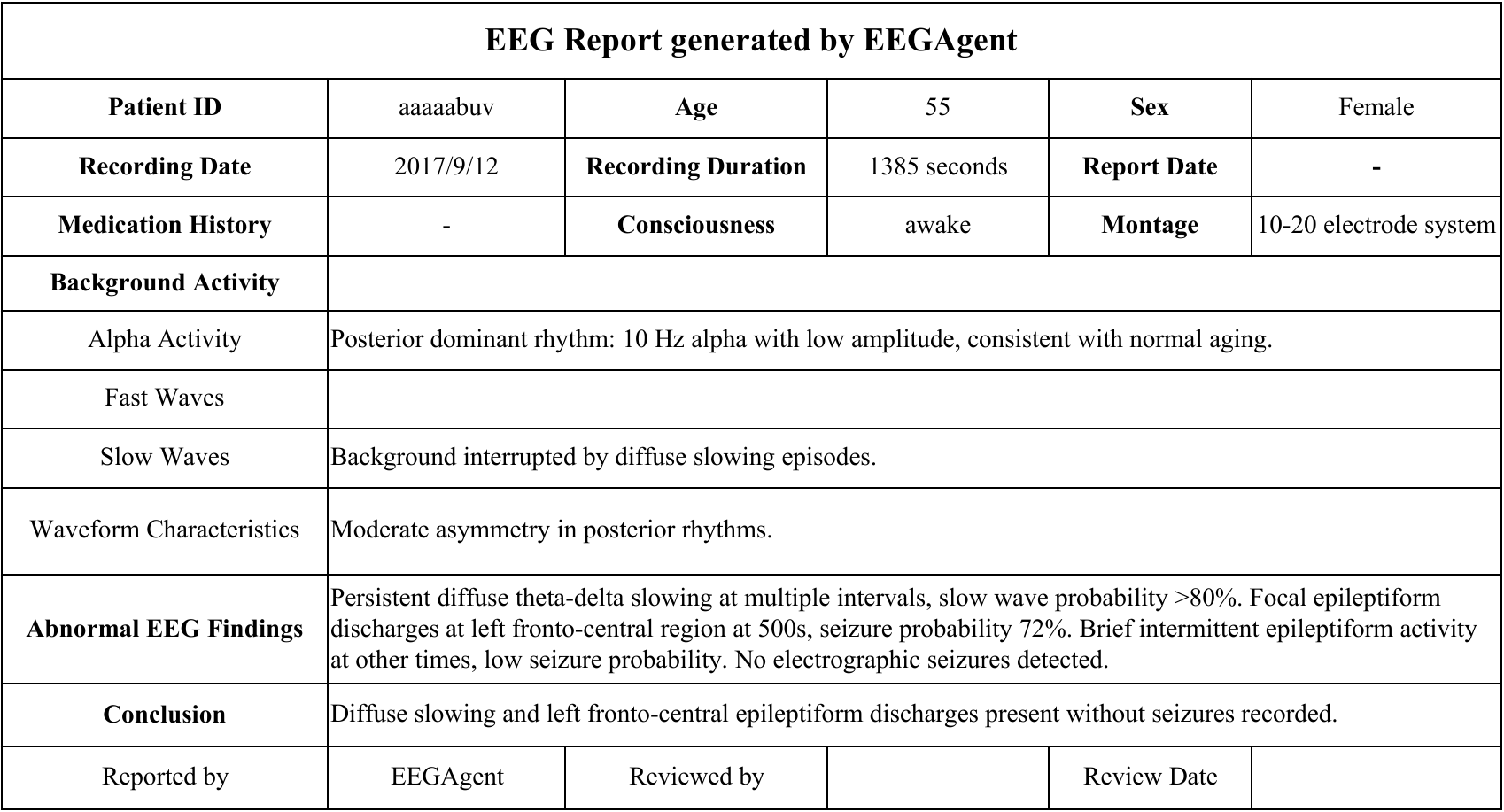} 
	\caption{EEG report generated by the EEGAgent} 
	\label{fig:report_}
\end{figure}

In this task, the EEGAgent receives a full EEG recording and generates a structured report containing basic information, background activity, abnormal events, and diagnostic conclusions. 
As shown in Figure~\ref{fig:report_}, the Agent accurately extracts patient metadata (e.g., age, recording duration, electrode configuration), identifies background rhythms and their spatial distribution, and detects abnormal events at a coarse level. 
For example, in the illustrated case, the system described occipital alpha slowing, diffuse slow waves, and left fronto-central epileptiform discharges, and synthesized these findings into a clinically coherent conclusion. 
This experiment demonstrates that \textbf{the EEGAgent is capable of generating clinically interpretable EEG reports, covering key information elements and supporting abnormality detection and localization.} 
Although quantitative evaluation remains challenging, the TUAB-based validation suggests that the proposed approach is feasible for automated EEG interpretation.

\section{Conclusion}
In this work, we present EEGAgent, the first framework to integrate intelligent agent systems with EEG analysis. To handle the long duration and non-stationary nature of EEG signals, we design a flexible toolbox operating across multiple temporal and spatial granularities for adaptive and context-aware analysis. We showcase its core capabilities—perception, exploration, detection, interaction and report generation—and evaluate it on public datasets. EEGAgent improves analysis efficiency, scalability, and interpretability, highlighting the potential of LLM-driven agents to automate EEG interpretation and support clinical decision-making.
	
\bibliography{aaai2026}
	
\end{document}